%% file: main.tex
\documentclass[journal]{IEEEtai}

\usepackage[colorlinks,urlcolor=blue,linkcolor=blue,citecolor=blue]{hyperref}

\usepackage{color,array}
\usepackage{float}
\usepackage{graphicx}
\usepackage{amsmath}
\usepackage{amssymb}

\begin{document}

\title{DCRMTA: Deep Causal Representation Learning for Multi-Touch Attribution}

\author{
Jiaming Tang,
Jingxuan Wen,
Liping Jing
\thanks{Jiaming Tang is with the Department of Electrical and Computer Engineering, University of Michigan, Ann Arbor, MI, USA.}
\thanks{Jingxuan Wen and Liping Jing are with Beijing Jiaotong University, Beijing, China.}
}

\maketitle

\begin{abstract}
Multi-touch attribution (MTA) is essential for estimating the contribution of individual advertising touchpoints to user conversions. While recent studies incorporate causal inference to mitigate confounding bias from user preferences, existing multi-stage deconfounding pipelines exhibit a critical structural flaw: they indiscriminately filter out user influences, which inadvertently discards the genuine causal signals linking user covariates to conversions. To resolve this trade-off, we propose Deep Causal Representation for MTA (DCRMTA), an end-to-end framework that explicitly quantifies and preserves the causal impact of user features. By leveraging structural causal modeling and adaptive counterfactual attention perturbations, DCRMTA distills invariant user representations while actively decoupling them from latent confounding variables. Extensive experiments on real-world industrial datasets (Criteo and Taobao) demonstrate the efficacy of our approach. DCRMTA effectively improves predictive accuracy—yielding up to a 5.2\% relative improvement in PR-AUC over strong baselines—while providing robust, Shapley-based credit allocations across complex marketing channels.
\end{abstract}


\section{Introduction}
\IEEEPARstart{O}{nline} media platforms leverage various marketing channels, including social networks, video media, and push notifications, to deliver advertisements for advertisers. As the internet economy evolves, users' purchasing behaviors have grown increasingly complex\cite{richardson2007predicting}. Accurately quantifying the precise value of specific user interactions (e.g., ad views, clicks, or purchases) is critical for optimizing marketing budget allocation across channels. This attribution process evaluates the impact of online advertising campaigns on conversion rates, thereby facilitating the acquisition of new target demographics. Furthermore, with the growing recognition of the transformative impact of artificial intelligence (AI) across various domains, an increasing number of studies are leveraging AI to address the intricate dependency issues associated with marketing data.

As a prominent analytical framework in this domain, multi-Touch Attribution (MTA), also referred to as multi-campaign attribution, captures the underlying patterns of user interactions with advertising campaigns \cite{anderl2016mapping}. By leveraging historical data \cite{berman2018beyond}, MTA quantitatively assesses the contribution of individual marketing touchpoints to user conversions. Consequently, this mechanism empowers advertisers to identify the core drivers of user acquisition and significantly enhances the precision of ad placement.

\begin{figure}[t]
    \centering
        \centering
        \includegraphics[width=0.45\textwidth]{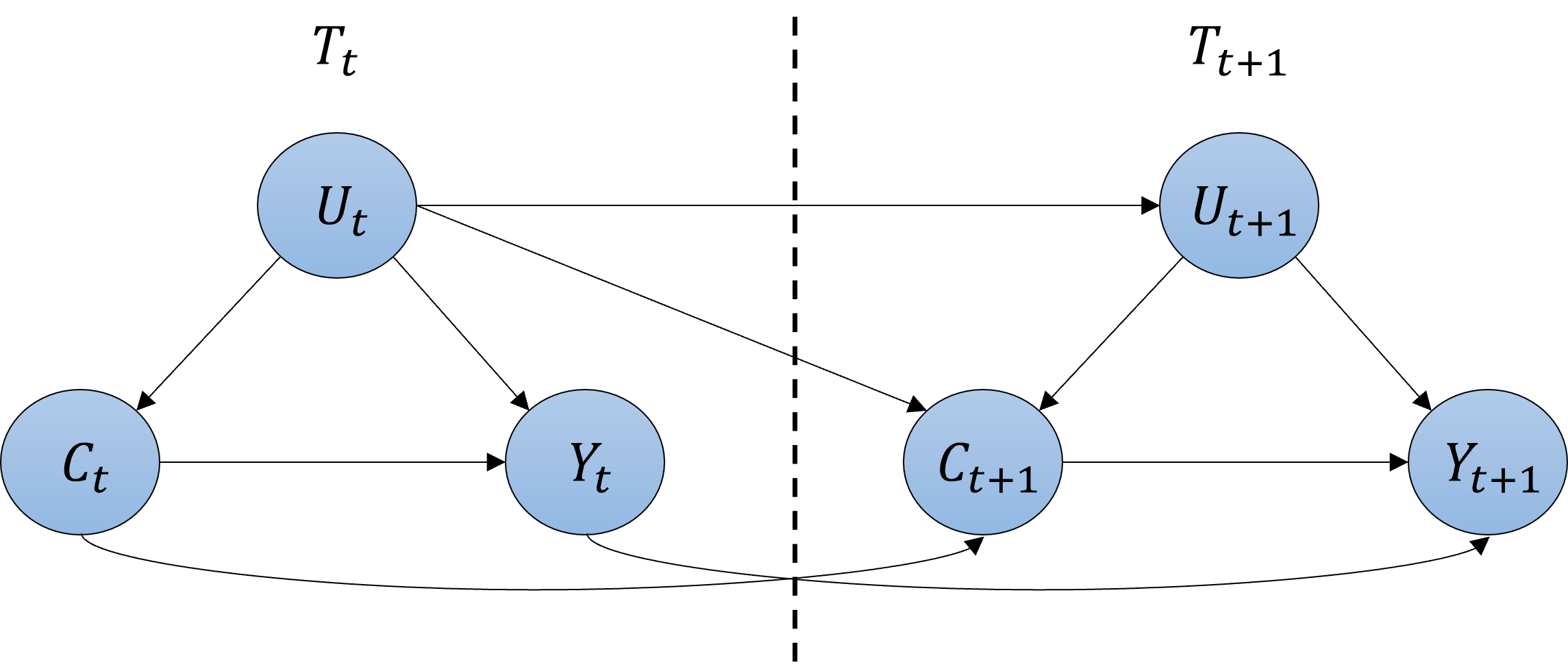} 
        \caption{Cause-and-effect diagram of touch point sequences which shows the causal model of touch point sequences. Representing the interaction between user $U$, advertising campaign $C$, and conversion result $Y$ at different times $T$.}
        \label{fig:subfig1}
   
\end{figure}
\begin{figure}[t]
        \centering
        \includegraphics[width=0.45\textwidth]{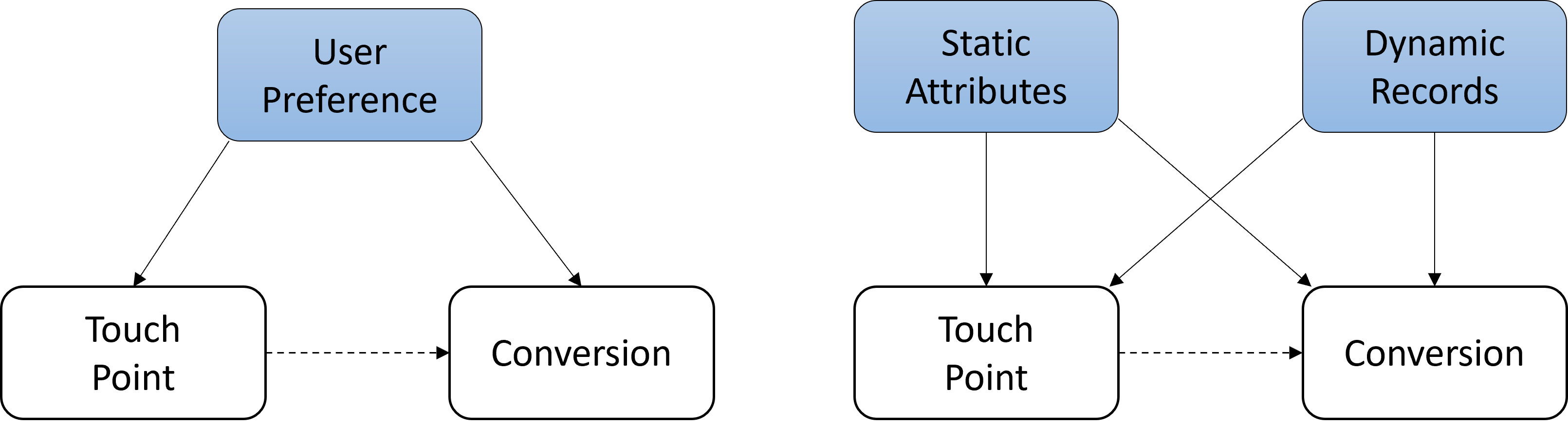} %
        \caption{Cause-and-effect diagram of user preferences. It shows the structure of user preference\cite{yao2022causalmta}. The left section represents the original diagram, while the right section differentiates user preferences into static attributes and dynamic records.}
        \label{fig:subfig2}
\end{figure}

Due to the intricate causal dependencies among data features, Multi-Touch Attribution (MTA) faces unique challenges. As illustrated in Figure \ref{fig:subfig1}, user features at time $T_t$ not only influence the current ad channels and conversion outcomes but also impact subsequent user states and channel exposures at time $T_{t+1}$. This phenomenon is primarily driven by the continuous interplay between online recommender systems and user behavior. Furthermore, this challenge is exacerbated by the fundamental problem of causal inference: the inability to simultaneously observe both positive ($Y_t(1)$) and negative ($Y_t(0)$) conversion outcomes. Consequently, estimating the true causal effect is impossible without counterfactual inference \cite{shalit2017estimating}. Therefore, accurately representing and distilling the causal relationships among users, channels, and conversions remains critical for addressing the MTA problem.

While the rapid advancements in deep learning have revolutionized various fields—including natural language processing, computer vision, and recommender systems—by demonstrating exceptional data representation capabilities, inherent complexities in MTA introduce significant performance bottlenecks. Although deep learning effectively extracts user attributes and ad touchpoint details, formidable challenges persist due to sparse touchpoint data, user selection bias \cite{chen2021bias}, and the missing-not-at-random (MNAR) problem \cite{rubin1976inference, marlin2009collaborative} arising from mutually exclusive ad exposures. Specifically, algorithms invariably tend to push advertisements toward individuals who are "naturally prone to conversion." This results in severe selection bias within the observational data we collect—a phenomenon that constitutes the core mechanism driving MNAR and hindering unbiased learning. As a result, user characteristics cannot be unbiasedly inferred from purely observational data. These conditions violate the underlying assumption of conversion models—unbiasedness—which assumes an equal likelihood of observing factual and counterfactual sequences, an assumption that rarely holds in real-world online marketing. Consequently, enhancing model performance necessitates the rigorous control of confounding variables to prevent biased estimations. To systematically address confounders in MTA, as shown in Figure \ref{fig:subfig2}, previous research \cite{yao2022causalmta} proposed categorizing user preference features into static attributes and dynamic records. Static attributes encompass demographic information such as age and gender, whereas dynamic records capture temporal behaviors, including historical browsing logs and media consumption patterns over a specific period.

Causal-based MTA presents an alternative paradigm to better explore and understand the underlying causal relationships between heterogeneous user touchpoint distributions and conversion outcomes. Integrating the robust representation capabilities of deep learning with the rigorous theoretical foundations of causal inference yields enhanced interpretability in the credit attribution domain. However, in practical scenarios, the full spectrum of user features cannot be comprehensively inferred relying solely on observational samples \cite{steck2011item}. Furthermore, the substantial influence of subjective factors on individual conversion behaviors inherently stems from variations in static attributes. Consequently, indiscriminately eliminating the impact of all static user features to adjust for confounding bias inadvertently restricts the model's ability to learn genuine user-campaign interaction patterns.

To overcome this representation dilemma, we propose DCRMTA, a novel deep causal representation learning framework tailored for the MTA problem. This approach effectively mitigates the confounding influence of static user features on conversion predictions while explicitly isolating variables that are causally related to conversions. Specifically, we design a Causal Attention Module(CAM) to distill distinct static representations from user-ad interaction histories. Furthermore, DCRMTA employs a counterfactual time series model based on Long Short-Term Memory (LSTM) \cite{hochreiter1997long} to process dynamic features, thereby predicting conversion probabilities through both individual characteristics and temporal user-ad interactions. Finally, the framework incorporates cooperative game theory to quantify the attribution weight of each ad touchpoint, facilitating optimal budget allocation and maximizing advertising ROI (Return on Investment).

\begin{figure*}[t]
    \centering
    \includegraphics[width=0.8\textwidth]{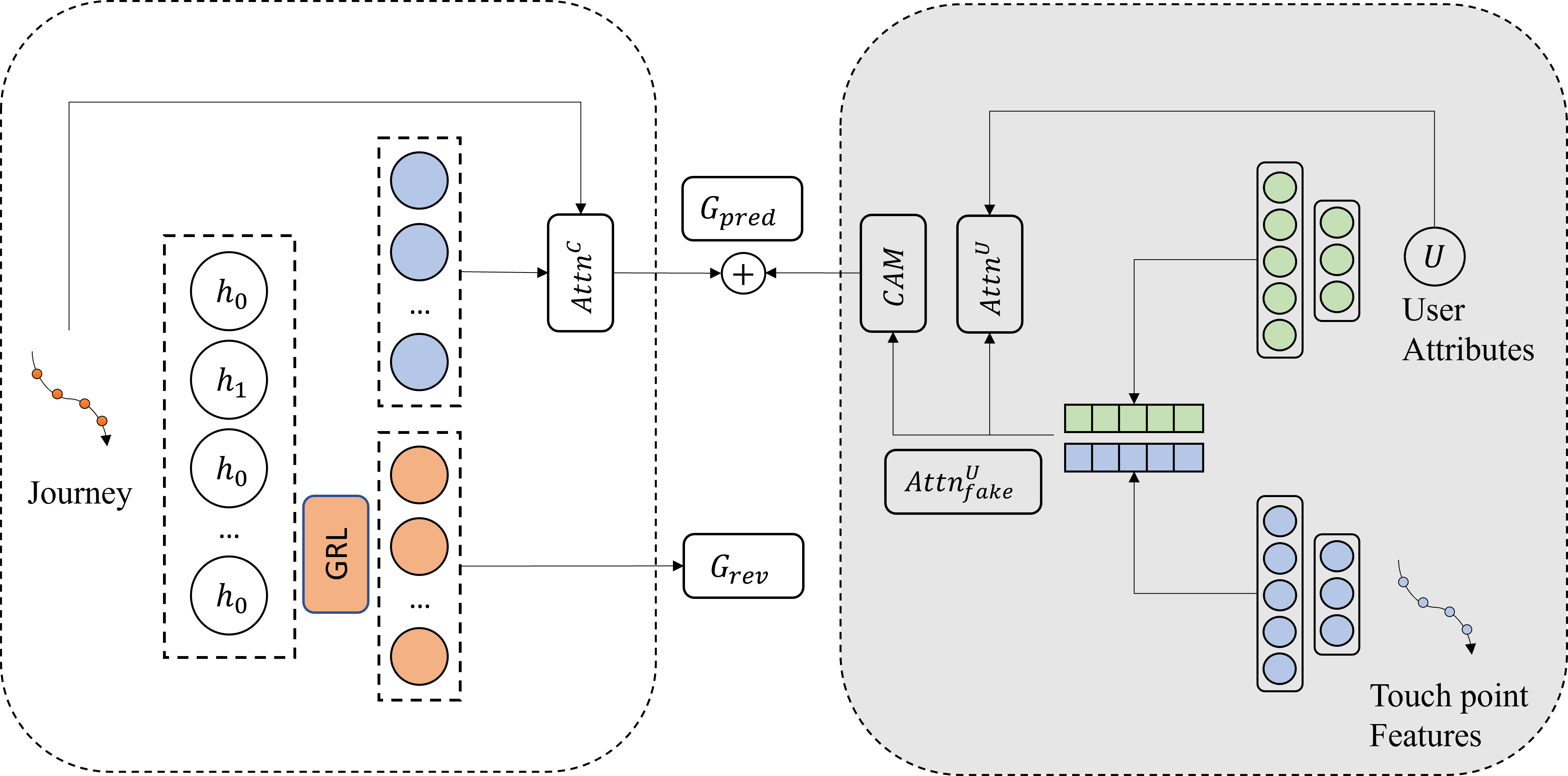}
    \caption{DCRMTA Architecture. GRL represents the gradient reversal layer \cite{bica2020estimating}, $Attn^C$ represents the attention layer of the advertising sequence, and $Attn^U$ represents the attention layer aggregated by the user and the touch feature sequence. CAM represents the Causal Attention Module, which is a pseudo attention mapping map $Attn^U_{fake}$ generated through zeroing and random insertion that implements counterfactual reasoning and calculates the causal effect of attention parameters that differ from the original reasoning result. $G_{pred}$ represents a binary classifier for conversion prediction, and $G_{rev}$ represents an advertising campaign classifier generated by gradient inversion.}
    \label{fig:overview}
\end{figure*}

The main contributions of this work are summarized as follows:
\begin{itemize}
\item End-to-end Causal Framework: To the best of our knowledge, we are the first to propose an end-to-end causal representation learning framework for the MTA task. As illustrated in Figure \ref{fig:overview}, DCRMTA leverages deep representation learning to automatically quantify complex causal effects among multifaceted user features.
\item Causal Attention Module: We design a novel Causal Attention Module that implements causal input perturbation on high-dimensional features, thereby achieving a more granular and refined causal effect estimation.
\item State-of-the-art Performance: Extensive experiments conducted on two large-scale real-world datasets (Criteo\footnote{https://ailab.criteo.com/criteo-attribution-modeling-bidding-dataset/} and Taobao\footnote{https://www.kaggle.com/datasets/pavansanagapati/ad-displayclick-data-on-taobaocom}) empirically validate the superior performance and robustness of our proposed method against state-of-the-art baselines.
\end{itemize}

\section{Related Works}
Existing literature on attribution methods can be broadly categorized into three paradigms: heuristic rules, data-driven MTA, and counterfactual prediction.

\textbf{Heuristic Rules.} Traditionally, market researchers relied on predefined, simplistic rules to allocate credit across advertising touchpoints, utilizing strategies such as last-click \cite{scaria2014last} and time decay \cite{wang2017display}. However, these intuitive approaches fundamentally disregard the heterogeneous characteristics of and temporal dependencies among various advertising channels. Consequently, their performance remains suboptimal as they fail to capture the complex causal dynamics inherent in user-campaign interactions.

\textbf{Data-Driven MTA.} To address these limitations, researchers shifted towards data-driven paradigms that learn attribution patterns directly from observational data, encompassing historical user profiles, ad exposure sequences, and conversion outcomes. Early data-driven MTA predominantly relied on probabilistic and statistical frameworks, subsequently incorporating survival analysis \cite{shao2011data} and hazard rate models \cite{zhang2014multi} to quantify the impact of ad exposure. Nevertheless, these conventional statistical methods largely neglect the influence of individual user characteristics on conversion probabilities and struggle to effectively model temporal inter-channel dependencies. Recently, leveraging the powerful sequential modeling capabilities of Recurrent Neural Networks (RNNs) \cite{rumelhart1986learning}, numerous deep learning-based MTA models have emerged to tackle these challenges. For instance, DNAMTA \cite{arava2018deep} integrates sophisticated feature encoding and a hierarchical attention mechanism to achieve competitive conversion prediction performance. DARNN \cite{ren2018learning} jointly models post-view and post-click attribution patterns to refine the final conversion estimation. Furthermore, DeepMTA \cite{yang2020interpretable} synergizes deep learning with cooperative game theory; it estimates the attribution contribution of each touchpoint by evaluating counterfactual sequences via the Shapley value, thereby yielding a highly interpretable attribution framework.

\textbf{Counterfactual Prediction.} Methods in this category formulate attribution strictly as a causal inference problem, relying heavily on the estimation of counterfactual outcomes. Early studies integrated classical statistical techniques—such as stratification, inverse probability weighting, and propensity score matching (PSM)—into MTA models to achieve unbiased individual-level causal estimations. More recently, researchers have infused counterfactual frameworks into Recurrent Neural Networks (RNNs). For instance, the Counterfactual Recurrent Network (CRN) \cite{bica2020estimating} constructs invariant representations of treatments (i.e., ad exposures) at each time step. This effectively decouples historical data features from treatment assignments, thereby adjusting for time-dependent confounders. Building upon CRN and hierarchical network architectures, CAMTA \cite{kumar2020camta} systematically addresses these time-varying confounding variables. Furthermore, CausalMTA \cite{yao2022causalmta}, the prior work most relevant to ours, successfully mitigates both static and dynamic confounding biases to train an unbiased conversion prediction model. However, to achieve this strict unbiasedness, it completely eliminates the entire influence of static user features, thereby inadvertently discarding the genuine causal signals these features naturally contribute to user conversions. This critical conceptual flaw forms the core intuition behind our proposed DCRMTA framework: rather than indiscriminately filtering out static attributes to achieve debiasing, DCRMTA is explicitly designed to disentangle and preserve their true causal effects, ensuring a more comprehensive and accurate attribution mechanism.

\section{Methodology}
\subsection{Preliminary}
To formalize the Multi-Touch Attribution (MTA) problem, we define a dataset $\mathcal{D}$ comprising $N$ user conversion journeys alongside the demographic information of $U$ unique users. For the $i$-th journey, we define the temporal sequence of touchpoints as $\{tp_{t}^{i}\}_{t=1}^{T_{i}}$, where $T_{i}$ denotes the sequence length. Additionally, let $\mathcal{C}$ represent the set of ad campaigns, encompassing $k$ distinct advertising channels.

Specifically, each user journey in dataset $\mathcal{D}$ is formulated as a tuple $(\mathcal{J}^{i}, y^{i})$. The trajectory $\mathcal{J}^{i}$ is represented as:
\begin{equation}
    \mathcal{J}^{i} = (\mathcal{C}^{i}, \mathcal{F}^{i}) = \{(c_{t}^{i}, f_{t}^{i})\}_{t=1}^{T_{i}} \quad 
\end{equation}
where $\mathcal{C}^{i}$ and $\mathcal{F}^{i}$ denote the sequence of exposed ad campaigns and the corresponding touchpoint features, respectively. Each touchpoint $tp_{t}^{i} = (c_{t}^{i}, f_{t}^{i})$ consists of a campaign index $c_{t}^{i} \in \mathcal{C}$ indicating the specific channel exposed, and a feature vector $f_{t}^{i}$ capturing the dynamic contextual information at that timestep. The conversion outcome is denoted by a binary indicator $y^{i} \in \{0, 1\}$, representing whether the $i$-th journey ultimately culminated in a successful conversion.

The primary objective of MTA is to model these sequential patterns and allocate precise attribution credits to all constituent touchpoints $\{tp_{t}^{i}\}_{t=1}^{T_{i}}$. However, historical observational datasets are inherently plagued by temporal dependencies and confounding biases stemming from user preferences. Consequently, standard deep learning models, which merely capture statistical correlations, fail to eliminate the catastrophic impact of these confounders on prediction accuracy. To address this, causal representation learning for MTA aims to distill the true causal patterns of users and campaigns, thereby facilitating the estimation of unbiased attribution credits.

\subsection{Journey causal feature extraction}

To model the sequential structure inherent in the channel time-series data, our objective is to eliminate the dynamic confounding variables induced by time-varying user features and recommendation mechanisms. Specifically, we adopt a hierarchical attention mechanism \cite{yang2016hierarchical} utilizing varying time windows to capture contextually relevant information via local and global attention maps. These extracted features are subsequently aggregated based on their respective importance weights to formulate robust predictive representations.

To explicitly mitigate dynamic bias, we introduce a Gradient Reversal Layer (GRL) to construct a sequence reconstruction module. This architecture compels the model to learn causal data representations through an adversarial training paradigm against the conversion prediction task. By maximizing the ad sequence generation loss while simultaneously minimizing the conversion prediction loss, our journey causal feature extraction module adversarially explores the generation of diverse campaign sequences from a shared underlying distribution while modeling the time-series dynamics. This component primarily consists of two sub-modules: an Attention Journey Model and a Reverse Generation Model.

\begin{figure}[t] 
    \centering
    \includegraphics[width=0.25\textwidth]{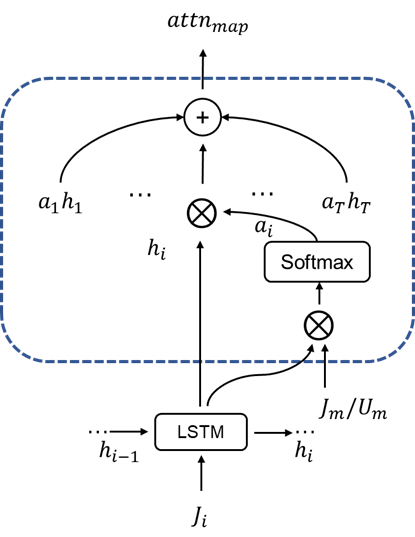}
    \caption{Hierarchical attention mechanism in DCRMTA}
    \label{fig:HAM}
\end{figure}
Merge vectors after word embedding encoding to obtain the input of the model:
\[
\mathrm{e_c}, \mathrm{e_{\mathcal{F}}} = \mathrm{Embedding}(\mathcal{C}, \mathcal{F})
\]
\[
\mathrm{e_{in}} = \mathrm{concat}(\mathrm{e_c}, \mathrm{e_{\mathcal{F}}})
\]

For embedded layer input $\{\mathrm{e_{in1}}, \mathrm{e_{in2}}, \ldots, \mathrm{e_{inT}}\}$, each LSTM cell based on the current input $\mathrm{e_{int}}$ and the previous hidden layer state output $\mathrm{h_{t-1}}$ to update the state of the current hidden layer $\mathrm{h_t}$ where $\mathrm{h_t} = \mathrm{h}(\mathrm{e_{in}}, \mathrm{h_{t-1}})$.

\textbf{Attention Journey module.} As illustrated in Figure \ref{fig:HAM}, we leverage the hidden states $\mathrm{h}$ generated by the LSTM layer to aggregate the core features of the campaign sequences via the hierarchical attention mechanism. Formally, this attentive aggregation is computed as follows:$$\mathrm{h_t^{attn}} = \tanh(\mathrm{W_h} \mathrm{h_t} + \mathrm{b_h})$$$$\mathrm{a_t} = \frac{\exp((\mathrm{h_t^{attn}})^\top \mathrm{e_{in}})}{\sum_{j} \exp((\mathrm{h_j^{attn}})^\top \mathrm{e_{in}})}$$$$\mathrm{m} = \sum_t \mathrm{a_t} \mathrm{h_t}$$where $\mathrm{a_t}$ denotes the attention weight assigned to the $t$-th hidden state $\mathrm{h_t}$, and $\mathrm{m}$ represents the extracted causal representation of the advertising sequence. This aggregated vector $\mathrm{m}$ subsequently serves as the input for the fusion conversion prediction module in section \ref{sec: Fusionconversion}.

\textbf{Reverse Generation Module.} In this module, we integrate a Gradient Reversal Layer (GRL) with a Multi-Layer Perceptron (MLP) to perform counterfactual classification of advertising channels. The predicted channel distribution $\mathrm{c_t^{rev}}$ is obtained via:
\[
\mathrm{h_t^{rev}} = \mathrm{MLP}(\mathrm{GRL}(\mathrm{h_t}))
\]
\[
\mathrm{c_t^{rev}} = \mathrm{softmax}(\mathrm{h_t^{rev}})
\]
where $\sigma$ represents the activation function.

Based on the generated predictions $\mathrm{c_t^{rev}}$, the sequence generation loss $\mathcal{L}_{\mathrm{rev}}$ is quantified using the Cross-Entropy (CE) objective:
\begin{equation}
    \mathcal{L}_{\mathrm{rev}} = \alpha \sum_{i=1}^N \sum_{t=1}^{T_i} \mathrm{CE}(\mathrm{c_t^{rev}}, \mathrm{c_t})
\end{equation}
where $\mathrm{CE}(\cdot, \cdot)$ represents the standard cross-entropy loss function, and $\alpha$ is a tunable hyperparameter controlling the strength of the adversarial penalty.

\subsection{User Causal Feature Extraction}
In MTA, user interaction data intrinsically comprises both causal signals and confounding factors. Consequently, it is important to isolate the invariant features that consistently drive conversion outcomes. Within the causal representation learning paradigm, observational data is typically conceptualized as being generated by invariant "content" and trivial "style" variables, where only the content representations intrinsically dictate downstream predictions independent of stylistic shifts. Leveraging this perspective, we investigate the causal mechanism of attention by actively intervening on attention weights while preserving the underlying content representations. Specifically, by synthesizing counterfactual attention distributions and executing causal interventions, we evaluate the discrepancies in prediction outcomes before and after the intervention to distill the invariant causal effects encapsulated within user features. To guarantee representation stability during this process, Gaussian-based perturbations are introduced; this ensures the counterfactual attention distribution remains bounded near the original observational distribution, thereby preventing the distortion of semantic content. Driven by these theoretical insights, we propose a novel Causal Attention Module (CAM) to explicitly disentangle user causal features and quantitatively assess their true contributions to conversion forecasting. For a comprehensive theoretical derivation, please refer to Appendix \ref{appendix: A}.

\begin{figure}[t]
        \centering
        \includegraphics[width=0.35\textwidth]{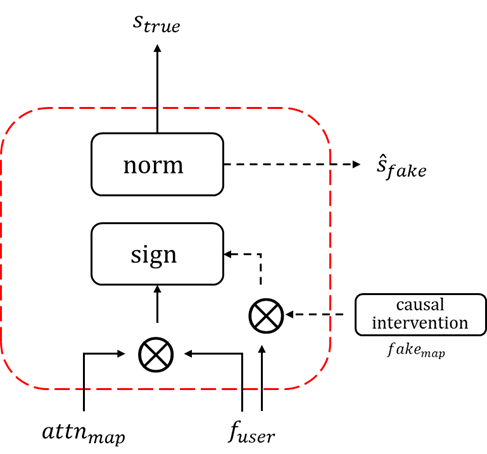}
        \caption{CAM module structure in DCRMTA}
        \label{fig:CAM}
\end{figure}

\textbf{Causal Attention Module.} In MTA, we introduce CAM to distill invariant causal representations linking user features and conversion outcomes. Specifically, this is achieved by automatically generating counterfactual attention distributions (i.e., "fake" attention maps) to simulate causal interventions, allowing us to explicitly quantify the isolated causal effect on conversion prediction. The detailed architecture of this module is illustrated in Figure \ref{fig:CAM}.

Formally, let the $i$-th user journey in dataset $\mathcal{D}$ be represented as a tuple $(\mathcal{U}^i, \mathcal{F}^i, y^i)$, where $\mathcal{U}^i$ denotes the static user attributes and $\mathcal{F}^i$ represents the sequence of interactive touchpoint features. These inputs are first projected into a dense vector space via embedding layers:
\[
\mathrm{e_u, e_F} = \mathrm{Embedding}(\mathcal{U}^i, \mathcal{F})
\]
Subsequently, non-linear transformations are applied to derive their latent representations:
\[
\mathrm{v_u} = \sigma (\mathrm{W_u} \cdot \mathrm{e_u} + \mathrm{b_u})
\]
\[
\mathrm{v_F} = \sigma (\mathrm{W_F} \cdot \mathrm{e_F} + \mathrm{b_F})
\]
\[
\mathrm{v_c} = \mathrm{concat}(\mathrm{v_u}, \mathrm{v_F})
\]
where $\mathbf{W}_*$ and $\mathbf{b}_*$ represent the learnable weight matrices and bias terms of the fully connected layers, respectively, and $\sigma$ denotes the activation function. Simultaneously, the contextual similarity between the user representations and the advertising sequence features is computed via matrix multiplication and normalized to form the attention weights $\mathbf{w}$:

$$w_t = \frac{\exp \left( (\mathbf{v}_{F, t})^\top \mathbf{v}_u \right)}{\sum_{j} \exp \left( (\mathbf{v}_{F, j})^\top \mathbf{v}_u \right)}$$
The attention-enhanced representation of the user touchpoint sequence is then aggregated as:

\[
\mathrm{v_{attn}} = (\mathrm{v_F})^\top \mathrm{w}
\]

To extract the final factual representation mapping user touchpoints to conversions, we perform a global aggregation operation. The concatenated feature $\mathbf{v}_c$ and the attention-enhanced feature $\mathbf{v}_{\text{attn}}$ are processed through an activation function $\psi$ (e.g., global average pooling) and subsequently normalized:

\[
\mathrm{s_i} = \psi (\mathrm{v_c}, \mathrm{v_{attn_i}}) = \frac{1}{HW} \sum_{h=1}^{H} \sum_{w=1}^{W} \mathrm{v_c^{h,w}} \mathrm{v_{attn_i}^{h,w}}
\]
\[
\mathrm{s} = \mathrm{norm}([\mathrm{s_1}, \mathrm{s_2}, \ldots, \mathrm{s_M}])
\]
where $\mathrm{s_*}$ is the output of sign activation function.

NNext, to perform a causal intervention, we construct a counterfactual representation $\mathbf{v}_{\text{fake}}$ by replacing the learned attention weights with an initialization matrix drawn from a Gaussian noise distribution. We apply the identical aggregation operation to derive the counterfactual state:
\[
\mathrm{\hat{s_i}} = \psi (\mathrm{v_c}, \mathrm{v_{fake_i}}) = \frac{1}{HW} \sum_{h=1}^{H} \sum_{w=1}^{W} \mathrm{v_c^{h,w}} \mathrm{v_{fake_i}^{h,w}}
\]
\[
\mathrm{\hat{s}} = \mathrm{norm}([\mathrm{\hat{s}_1}, \mathrm{\hat{s}_2}, \ldots, \mathrm{\hat{s}_M}])
\]

\subsection{Fusion conversion prediction}
\label{sec: Fusionconversion}
By integrating the outputs from the aforementioned modules, we fuse the causal representation of the advertising sequence ($\mathbf{m}$) with the user causal features ($\mathbf{s}$) to perform the final conversion prediction via an MLP-based binary classifier:
\[
\mathrm{v}_\mathrm{final} = \mathrm{s} + \mathrm{m}
\]
\[
\mathrm{p} = \sigma (\mathrm{MLP}(\mathrm{v}_\mathrm{final}))
\]
where $p$ represents the predicted conversion probability. To enforce the causal constraint, we isolate the direct causal effect by contrasting the factual predictions with the counterfactual predictions:
\[
\hat{\mathrm{v}}_\mathrm{final} = \hat{\mathrm{s}} + \mathrm{m}
\]
\[
\mathrm{v}_\mathrm{eff} = \sigma(\mathrm{v}_\mathrm{final}) - \sigma(\hat{\mathrm{v}}_\mathrm{final})
\]
where $\mathbf{v}_{\text{eff}}$ captures the distinct causal effect. The objective function for the causal conversion prediction module consists of two components: the standard conversion prediction loss and the causal effect penalty:
\begin{equation} 
    \mathcal{L}_{\mathrm{cpred}} = \beta \sum_{i=1}^{N} \mathrm{CE}(\mathrm{p_i}, y_i) + \gamma \sum_{i=1}^{N} \mathrm{MSE}(\mathrm{v}_\mathrm{eff}, \textbf{0}) 
\end{equation}

where $\text{CE}(\cdot, \cdot)$ is the cross-entropy loss, $\text{MSE}(\cdot, \cdot)$ denotes the mean squared error.

Ultimately, the overall objective function of the proposed DCRMTA framework jointly optimizes the sequence reconstruction loss and the causal conversion prediction loss, formulated as:
\begin{equation} \label{eq5}
    \mathcal{L}_\mathrm{total} = \mathcal{L}_\mathrm{cpred} + \mathcal{L}_{\mathrm{rev}} 
\end{equation}

Here, $\alpha$, $\beta$, and $\gamma$ serve as scaling hyperparameters to balance the multi-task learning objectives.

\subsection{Credit Attribution}
For attribution credits allocation of ads, DCRMTA uses Shapley Values\cite{shapley1953value}, which has been widely used.

First, we manually create counterfactual ad journeys by deleting touchpoints $tp_t^i$. $\mathcal{S}$ denotes a subsequence of counterfactual ad journey $\mathcal{S} \subseteq \mathcal{J}^i \setminus \{tp_t^i\}$. We use $\mathrm{f}$ to define DCRMTA conversion model as a function, which gets ad journey as input and gives the conversion probability. Then the Shapley Values for a single ad campaign $c_t^i$ can be defined as:
\begin{equation}
    \label{eq7}
{SV_t^i = \sum_{\substack{S \subseteq \mathcal{J}^i \setminus \{t_{pi}\}}} 
\frac{|S|! \, (|\mathcal{J}^i| - |S| - 1)!}{|\mathcal{J}^i|!} \left[ f(S \cup \{c_i^t\}) - f(S) \right]}
\end{equation}

Afterwards, we use normalized Shapley Values to allocate attribution credits.
\begin{equation} \label{eq7}
\mathrm{attr_t^i} = \frac{(\mathrm{SV_t^i})^+}{\sum_{t=1}^{T_i} (\mathrm{SV_t^i})^+}
\end{equation}
Where $(\mathrm{x})^+ = \max (\mathrm{x}, 0)$, and $\mathrm{attr_t^i}$ are attribution credits for each campaign of $\mathcal{J}^i$ in time spot $\mathrm{t}$.

\section{Experiments}
In this section, we conduct extensive experiments to evaluate the performance of the proposed DCRMTA framework. Our evaluation aims to address the following four research questions (RQs):

RQ1: Predictive Efficacy. How does DCRMTA perform compared to state-of-the-art (SOTA) multi-touch attribution methods in terms of conversion prediction accuracy?

RQ2: Attribution Robustness. To what extent does the proposed model yield reliable and interpretable attribution results in complex, real-world marketing scenarios?

RQ3: Module Interpretability. What are the specific contributions and internal mechanisms of the Journey Causal Feature Extraction and User Causal Feature Extraction modules in capturing high-level representations?

RQ4: Causal Debiasing. How effective is DCRMTA in mitigating confounding bias and identifying authentic causal features?

\subsection{Experimental Settings}

\textbf{Dataset.} To evaluate the efficacy of the proposed DCRMTA framework, we conduct experiments on two large-scale real-world datasets. The statistical summaries are provided in Table \ref{tab:dataset}. 
(i) Criteo-Processed: Following the experimental protocols established in existing MTA literature, we first utilize the Criteo-Raw dataset, a benchmark widely recognized for online advertising click-through rate (CTR) prediction. To adapt it for the MTA task, we re-encode the feature fields and aggregate raw logs into user-centric sequences based on unique identifiers. Each touchpoint within an advertising campaign comprises nine categorical feature fields. Specifically, we follow the preprocessing methodology of prior studies to construct the Criteo-Processed dataset, ensuring a consistent and fair comparison with state-of-the-art baselines. (ii) Taobao: The second dataset consists of real-world e-commerce interaction logs from Taobao.com. This dataset encompasses user-ad interaction trajectories, advertisement metadata, and demographic profiles. To define conversion labels, journeys containing three or more clicks are categorized as potential conversions. For data quality assurance, we filter out campaigns with fewer than five total exposures and retain only the top 30 most frequent campaigns. For each user, touchpoints are sequenced chronologically and merged with corresponding ad/user attributes, resulting in a high-dimensional feature space inclusive of nine categorical attributes, campaign IDs, and temporal features.

\begin{table}[t]
\centering
\renewcommand{\arraystretch}{1.2}  
\caption{Statistics comparison between Criteo-Raw, Criteo-Processed, Taobao datasets.}
\resizebox{0.48\textwidth}{!}{
\begin{tabular}{lccc}
\hline
\textbf{Statistics} & \textbf{Criteo-Raw} & \textbf{Criteo-Processed} & \textbf{Taobao-Processed} \\ \hline
\textbf{\# users} & 6,142,256 & 157,331 & 101,731 \\
\textbf{\# campaign} & 675 & 10 & 30 \\ 
\textbf{\# journeys} & 6,514,319 & 196,560 & 101,731 \\ 
\textbf{\# convert journeys} & 435,810 & 19,890 & 50,544 \\ 
\textbf{\# touchpoints} & 16,468,027 & 787,483 & 924,765 \\ \hline
\label{tab:dataset}
\end{tabular}}
\end{table}

\begin{table*}[t]
\centering
\caption{Model Performance Comparison (\%)}
\renewcommand{\arraystretch}{1.2}
\setlength{\tabcolsep}{5pt}
\begin{tabular}{l|cccccc|cccccc}
\hline
Dataset & \multicolumn{6}{c|}{Criteo} & \multicolumn{6}{c}{Taobao} \\
 Model & Acc. & Recall & Prec. & F1 & PR-AUC & LogLoss & Acc. & Recall & Prec. & F1 & PR-AUC & LogLoss \\
\hline
SP         & 80.33  & 46.93  & 27.02  & 34.30  & 27.87  & 50.34  & 59.46  & 65.84  & 58.57  & 61.99  & 62.38  & 68.31 \\
LR         & 83.23  & 35.87  & 28.30  & 31.64  & 30.55  & 70.47  & 50.75  & \textbf{99.01}  & 50.48  & 66.87  & 68.92  & 64.40 \\
DARNN      & 89.75  & 67.38  & 52.05  & 58.73  & 61.08  & 38.42  & 61.87  & 84.53  & 58.29  & 69.00  & 71.38  & 61.92 \\
DNAMTA     & 93.09  & 71.59  & 66.91  & 69.17  & 75.41  & 30.43  & 62.87  & 84.03  & 59.17  & 69.44  & 75.01  & 60.09 \\
CAMTA      & 91.78  & 71.15  & 60.15  & 65.19  & 73.42  & 31.74  & 61.51  & 86.53  & 57.79  & 69.30  & 74.83  & 60.36 \\
CausalMTA  & 92.78  & 71.73  & 65.09  & 68.25  & 74.26  & 31.87  & 62.86  & 83.43  & 59.24  & 69.29  & 74.87  & 60.37 \\
\hline
DCRMTA-w/o-CAM & 93.54  & \textbf{76.20}  & 67.96  & 71.84  & 77.97  & 28.54  & 62.48  & 84.92  & 58.74  & 69.44  & 75.07  & 60.14 \\
DCRMTA-w/o-U   & 93.19  & 76.08  & 66.10  & 70.74  & 77.54  & 29.34  & 62.21  & 85.09  & 58.50  & 69.33  & 74.77  & 60.32 \\
DCRMTA-w/o-J   & 89.98  & 72.31  & 52.70  & 60.97  & 54.95  & 37.08  & 60.87  & 86.98  & 57.26  & 69.06  & 74.68  & 60.49 \\
\textbf{DCRMTA}     & \textbf{93.73}  & 75.62  & \textbf{69.27}  & \textbf{72.30}  & \textbf{79.44}  & \textbf{28.31}  & \textbf{63.49}   & 84.87  & \textbf{60.04}  & \textbf{70.32}  & \textbf{75.17}  & \textbf{59.47} \\
\hline
\end{tabular}
\label{tab:model_perf_comp}
\end{table*}

\begin{table*}[t]
\centering
\caption{Results of data reply on Criteo-Processed dataset.}
\label{table:datareplay}
\setlength{\tabcolsep}{5pt}
\renewcommand{\arraystretch}{1.3}
\begin{tabular}{l|cccc|cccc|cccc}
\hline
Criterion  & \multicolumn{4}{c|}{CPA} & \multicolumn{4}{c|}{CR} & \multicolumn{4}{c}{CN} \\ 
Budget Prop.& 1/2 & 1/4 & 1/8 & 1/16 & 1/2 & 1/4 & 1/8 & 1/16 & 1/2 & 1/4 & 1/8 & 1/16 \\ \hline
DARNN       & 1427.80 & 1284.60 & 1156.89 & 1000.79 & 0.0795 & 0.0582 & 0.0434 & 0.0333 & 2426 & 1775 & 1323 & 1017 \\
CAMTA       & \textbf{1280.26} & 1135.68 & 982.06 & 841.45 & 0.1548 & 0.1057 & 0.0767 & 0.0569 & 4725 & 3225 & 2342 & 1736 \\
DNAMTA      & 1425.95 & 1267.24 & 1101.22 & 922.97 & \textbf{0.1740} & 0.1158 & 0.0801 & 0.0566 & \textbf{5310} & 3534 & 2444 & 1726 \\
CausalMTA   & 1299.38 & 1130.94 & 967.26 & 833.58 & 0.1622 & \textbf{0.1189} & 0.0866 & 0.0624 & 4951 & \textbf{3628} & 2643 & 1903 \\ \hline
DCRMTA  & 1281.14 & \textbf{1112.88} & \textbf{953.47} & \textbf{816.96} & 0.1585 & 0.1160 & \textbf{0.0871} & \textbf{0.0642} & 4838 & 3539 & \textbf{2658} & \textbf{1960} \\ \hline
\end{tabular}
\end{table*}

\textbf{Baselines.} To rigorously evaluate DCRMTA, we compare it against seven representative baseline methods and three ablation variants. These include traditional statistical approaches, SP \cite{dalessandro2012causally} and LR \cite{shao2011data}, as well as diverse deep learning architectures such as DeepMLP, DARNN \cite{zhou2018deep}, and DNAMTA \cite{arava2018deep}, where the latter utilizes deep neural networks to capture complex non-linear interactions within user journeys. Furthermore, we incorporate causal-based models, namely CAMTA \cite{kumar2020camta} and CausalMTA, to benchmark the framework's capability in mitigating spurious correlations via counterfactual learning and sequence reweighting. Finally, the internal contributions of our framework are validated through three ablation variants: DCRMTA-w/o-U, DCRMTA-w/o-J, and DCRMTA-w/o-CAM, which respectively exclude the User Causal Feature Extraction, Journey Causal Feature Extraction, and Causal Attention Modules.

\subsection{Model Performance Comparisons}

To address \textbf{RQ1}, we evaluate the efficacy of various models in conversion prediction. Table~\ref{tab:model_perf_comp} summarizes the comparative results on the Criteo and Taobao datasets, providing quantitative insights into how different architectures handle sequential conversion tasks. We employ a comprehensive suite of metrics—including Accuracy, Recall, Precision, F1-score, and Cross Entropy Loss—to assess predictive quality. Notably, we prioritize PR-AUC over ROC-AUC as our primary metric, as it offers a more rigorous assessment of the model’s discriminative power regarding positive samples (conversions) in the presence of class imbalance.

On the Criteo-Processed dataset, DARNN exhibits the most constrained performance among deep learning baselines, yielding a PR-AUC of 0.6108 and a Logloss of 0.3842. This marginal efficacy underscores the limitations of standard recurrent structures in capturing complex sequential interactions and latent causal dependencies. Conversely, more advanced frameworks like DNAMTA and CAMTA achieve substantial performance gains. CausalMTA further elevates the benchmark by integrating causal graph priors and structural constraints. The proposed DCRMTA achieves state-of-the-art (SOTA) performance, reaching a PR-AUC of 0.7944 and the lowest Logloss of 0.2831. This superior performance is attributed to the synergistic integration of counterfactual attention perturbation and explicit confounder disentanglement, which collectively fortify the model's causal inference capacity. Notably, DCRMTA achieves a 5.2\% relative improvement in PR-AUC over CausalMTA, validating its architectural advantages.

On the Taobao dataset, the performance gap between models narrows, primarily due to the inherent lack of user-level journey repeatability. Such data sparsity diminishes the discriminative signals typically exploited by attribution models. Nevertheless, DCRMTA maintains its lead, securing the highest F1-score and PR-AUC alongside the minimum Logloss. This result is particularly significant; while conventional causal methods often rely on recurring user behaviors to extract stable attribution signals, DCRMTA successfully learns invariant user representations through cross-user consistency. Even in the absence of intra-user repetitions, our framework’s attention perturbation and disentanglement mechanisms remain robust, effectively capturing generalized causal structures.

In summary, DCRMTA consistently outperforms all baselines across both datasets. The results provide strong empirical evidence for the efficacy of adaptive counterfactual perturbations and explicit confounder disentanglement. By accurately capturing individualized and structural causal effects, DCRMTA demonstrates the critical necessity of modeling causal structures within the multi-touch attribution pipeline.

\begin{figure}[h] 
    \centering
    \includegraphics[width=0.5\textwidth]{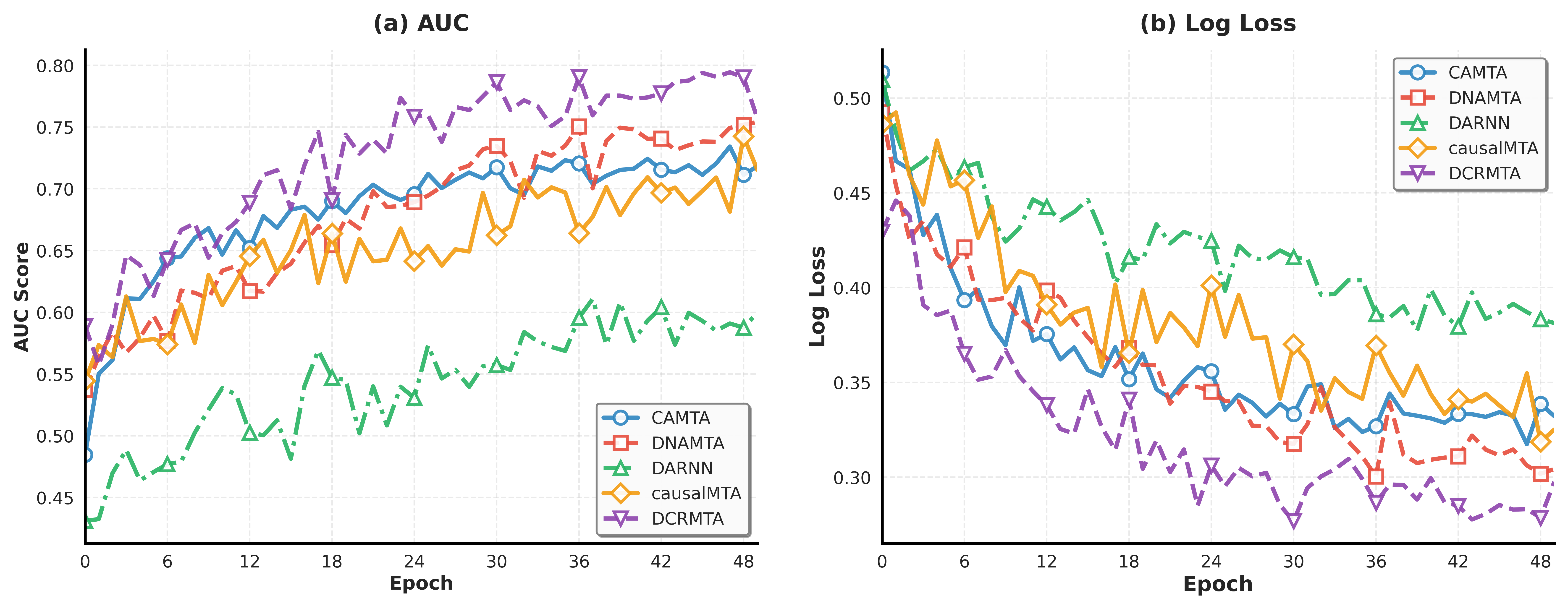}
    \caption{Convergence curves of deep MTA models on Criteo-Processed}
    \label{fig:convergence_curves}
\end{figure}

The training curves in Figure~\ref{fig:convergence_curves} not only demonstrate DCRMTA’s superior final performance, but also clearly highlight its faster convergence compared to all other models. In the early epochs (within the first 10 epochs), DCRMTA shows a sharp increase in PR-AUC and a rapid decline in LogLoss, indicating that it is able to learn effective representations and decision boundaries much more efficiently. While other models like DNAMTA and CausalMTA require over 30 epochs to approach their peak performance, DCRMTA reaches strong predictive accuracy and stability significantly earlier.

This accelerated convergence is primarily attributed to the integrated causal mechanisms—specifically intervention-based representation learning and explicit confounder disentanglement. By effectively isolating confounding variables and prioritizing invariant causal features, the model optimizes its parameters toward more generalizable representations from the onset of training. Consequently, DCRMTA attains robust performance with fewer iterations, underscoring its computational efficiency and practical viability for real-world deployment scenarios where training budgets are often constrained.

\subsection{Data Replay Evaluation}

To address \textbf{RQ2}, we perform data replay experiments on the Taobao dataset to evaluate the efficacy of various budget reallocation strategies in simulated real-world marketing environments. Following the evaluation protocols established in prior MTA research \cite{ren2018learning}, we employ three key performance indicators: Cost Per Action (CPA), Conversion Rate (CR), and Conversion Number (CN). Specifically, CN quantifies the total volume of successful conversions realized after the budget is reallocated according to the attribution scores generated by the respective models.

The experimental results, summarized in Table \ref{table:datareplay}, demonstrate the consistent superiority of DCRMTA across all metrics and budget scenarios. Regarding cost-efficiency, as the budget proportion is progressively constrained from $1/2$ to $1/16$, DCRMTA exhibits robust cost-control capabilities. Notably, under the most stringent budget constraint ($1/16$), DCRMTA achieves a CPA of 816.96, which is significantly lower than that of competing baselines. Furthermore, the CR analysis reveals that DCRMTA maintains the highest conversion efficiency—achieving 0.0871 and 0.0642 under the $1/8$ and $1/16$ budget levels, respectively—outperforming state-of-the-art (SOTA) methods such as CAMTA and CausalMTA. These results indicate that DCRMTA effectively optimizes marketing utility even in high-sparsity, budget-constrained environments.

Analysis of the CN metric further reinforces these findings. At lower budget proportions ($1/8$ to $1/16$), DCRMTA secures 2,658 and 1,960 successful conversions, respectively, validating its strategic advantage in maximizing budget utilization. Collectively, these empirical findings underscore the comprehensive benefits of the DCRMTA framework in enhancing cost-effectiveness and conversion yield, particularly in scenarios characterized by restricted marketing resources.

\subsection{Hyperparameter Sensitivity Analysis}
\begin{figure}[t]
  \centering
  \includegraphics[width=0.45\textwidth]{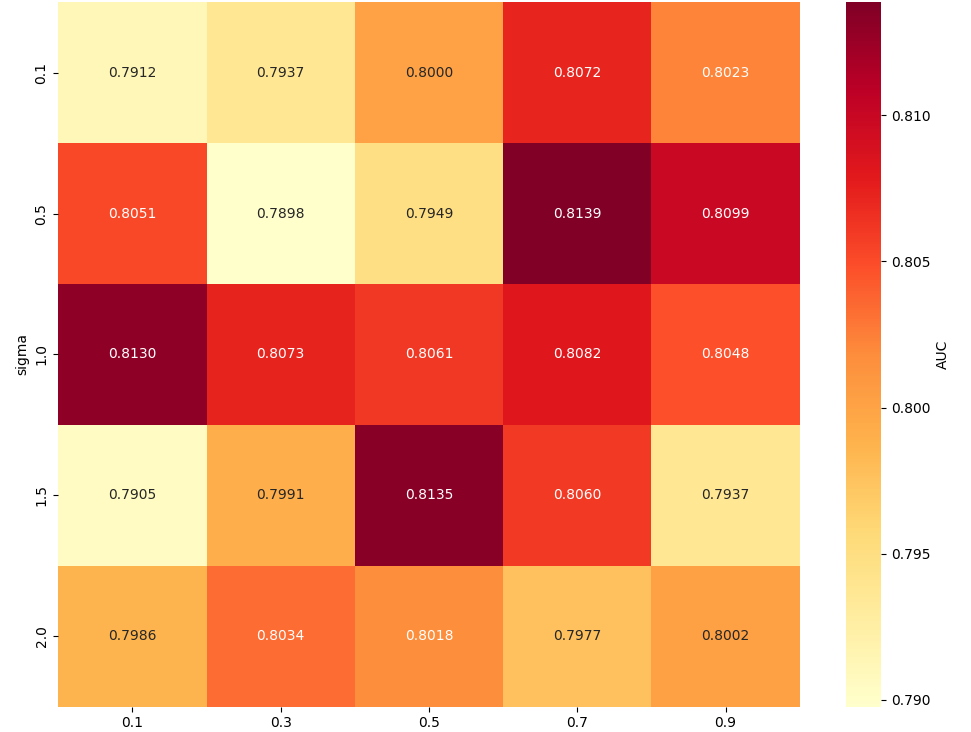}
  \caption{AUC heatmap showing the effect of counterfactual noise ($\sigma$) and loss weight ($\delta$).}
  \label{fig:auc_heatmap}
\end{figure}

\begin{figure}[t]
  \centering
  \includegraphics[width=0.45\textwidth]{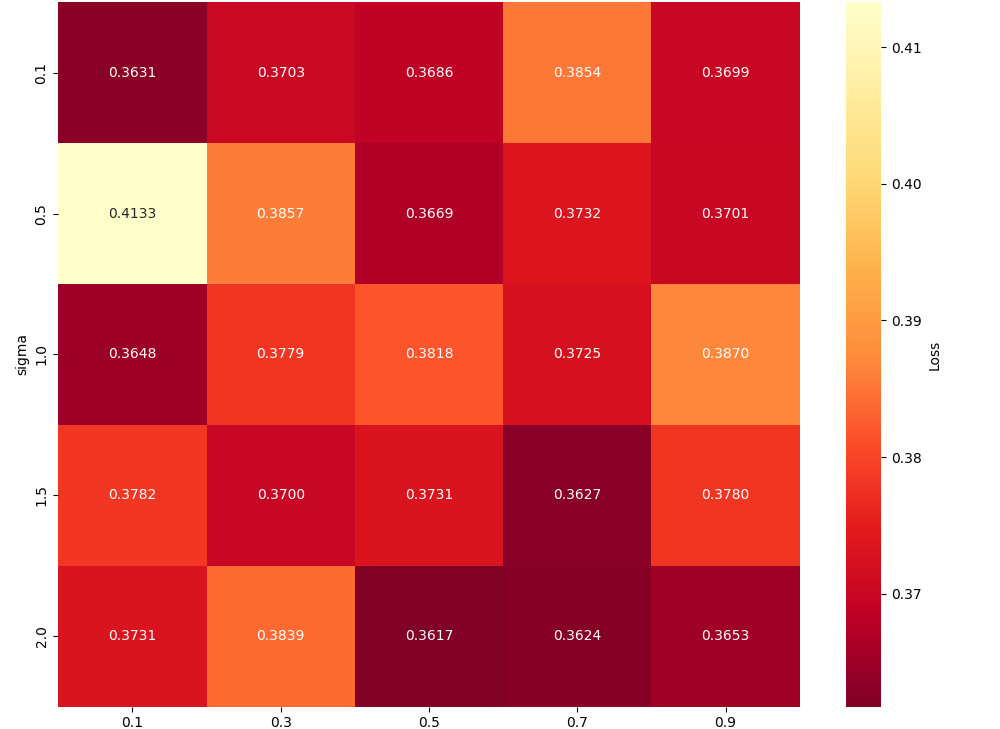}
  \caption{Loss heatmap showing the effect of counterfactual noise ($\sigma$) and loss weight ($\delta$).}
  \label{fig:loss_heatmap}
\end{figure}

To investigate the impact of counterfactual learning dynamics on DCRMTA, we perform a sensitivity analysis on the Criteo-Processed dataset, focusing on the noise standard deviation ($\sigma$) and the counterfactual loss weight ($\delta$). As illustrated in the heatmaps (Fig. \ref{fig:auc_heatmap} and \ref{fig:loss_heatmap}), the model achieves peak performance at $\sigma = 0.5$ and $\delta = 0.7$, yielding an AUC of 0.8139 and a minimal loss of 0.3627.

The empirical results reveal that model efficacy is highly sensitive to the noise scale $\sigma$. Specifically, increasing $\sigma$ from 0.1 to 1.0 consistently enhances the AUC, which plateaus within the $\sigma \in [0.5, 1.5]$ range before declining. This trend underscores that moderate stochastic perturbation is essential for generating informative counterfactual signals that facilitate robust causal representation learning. Conversely, the influence of the loss weight $\delta$ is more nuanced; its efficacy is strictly coupled with the selection of $\sigma$, where moderate $\delta$ values ($0.5 \le \delta \le 0.7$) yield optimal stability only when paired with sufficient noise levels.

In summary, the optimal configuration resides within a performance "ridge" defined by $\sigma \in [0.5, 1.5]$ and $\delta \in [0.5, 0.7]$. Given the high sensitivity to $\sigma$, these findings suggest that precise calibration or adaptive scheduling of counterfactual perturbations is critical for maximizing the framework's deconfounding capabilities.

\subsection{Ablation Studies}
\subsubsection{Features Extraction module ablation}
To address \textbf{RQ3}, we conduct a series of ablation experiments to quantify the individual contributions of the core components within the DCRMTA framework: the User Causal Feature Extraction (UCFE), the Journey Causal Feature Extraction (JCFE), and the Causal Attention Module(CAM). The comparative results are integrated into Table \ref{tab:model_perf_comp}.

\begin{figure*}[t]
  \centering
  \includegraphics[width=0.75\textwidth]{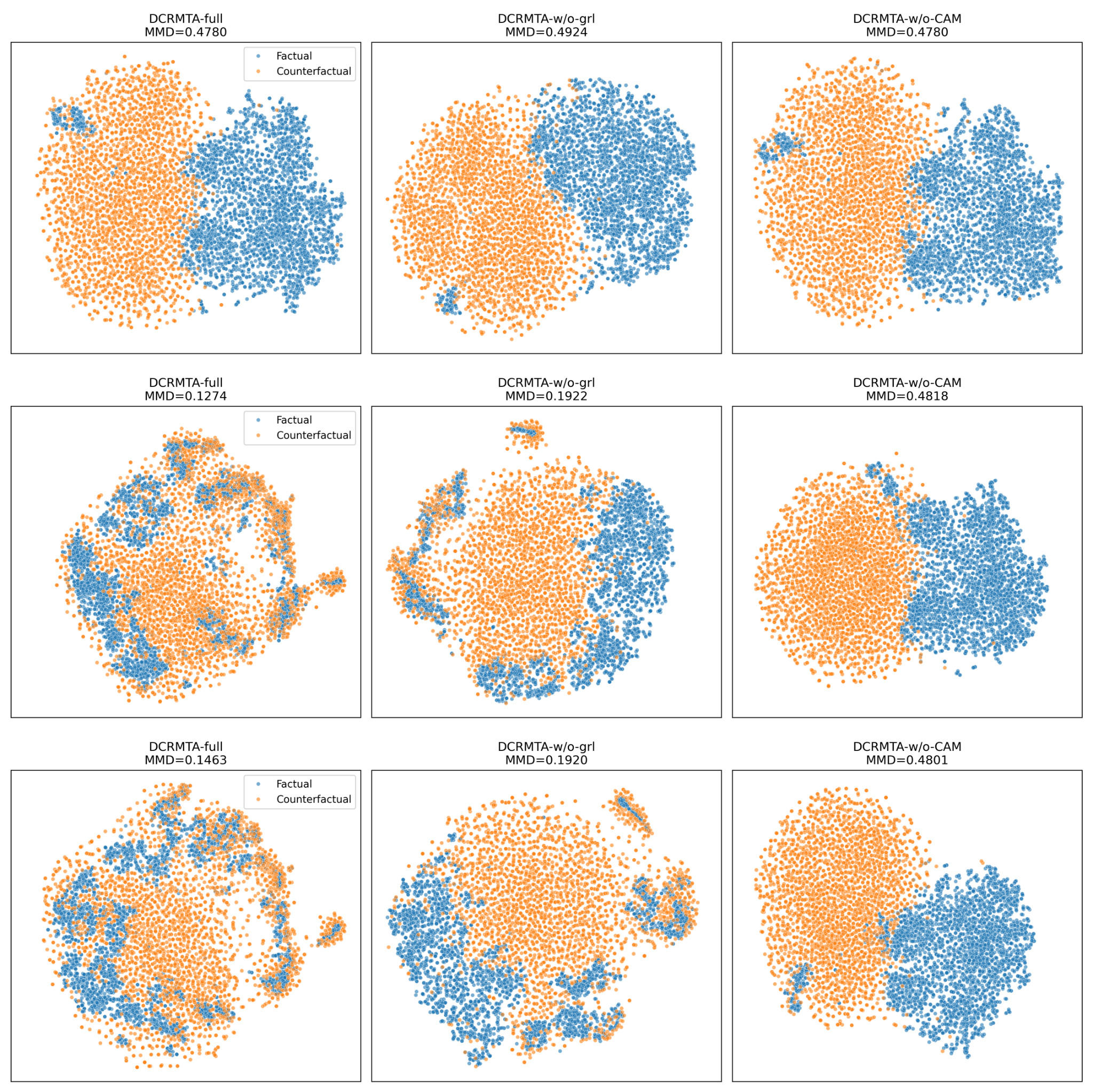}
  \caption{t-SNE visualization of learned representations for DCRMTA variants on the Criteo dataset. The blue and orange clusters represent factual and counterfactual distributions, respectively. Each row denotes different training stages (Top to Bottom: epoch 1, epoch 5, and epoch 10), while columns represent different model architectures. }
  \label{fig:var_stacked}
\end{figure*}

First, the removal of the JCFE and UCFE modules leads to significant performance degradation across both benchmarks, albeit to varying degrees depending on the dataset characteristics. On the Criteo dataset—characterized by rich campaign-level features and complex temporal dependencies—the exclusion of JCFE (DCRMTA-w/o-J) results in a substantial decline in PR-AUC (-24.49\%) and F1-score (-11.33\%). In contrast, the model without UCFE (DCRMTA-w/o-U) remains relatively stable, suggesting that the adversarial structure within the Journey module is the primary driver for extracting robust temporal causal representations in feature-rich environments. Conversely, on the Taobao dataset—where individual-level user attributes provide more discriminative signals—DCRMTA-w/o-U exhibits a more pronounced performance drop than DCRMTA-w/o-J. This shift underscores the effectiveness of the UCFE module in capturing fine-grained, individual-level causal characteristics when sequence-level repetition is sparse.

Second, the elimination of the CAM (DCRMTA-w/o-CAM) consistently marginalizes predictive accuracy on both datasets (e.g., a 1.47\% reduction in PR-AUC on Criteo). This consistent degradation validates that the Causal Attention Module successfully enhances the framework’s generalization by identifying invariant causal patterns within user features, effectively filtering out spurious correlations.

Overall, the full DCRMTA configuration achieves the superior performance on all metrics. These findings demonstrate the synergistic advantage of integrating user-centric, sequence-level, and causal-aware mechanisms into a unified deconfounding framework.

\subsubsection{Evolution Analysis of Attention Distribution}
To address \textbf{RQ4}, we evaluate the framework’s capability in mitigating confounding bias and extracting invariant causal features. We employ t-SNE to visualize the evolution of latent representations, specifically comparing factual and counterfactual latent vectors ($\mathbf{z}_f$ and $\mathbf{z}_c$) extracted from the penultimate layer. These pairs are generated via learned causal interventions within the CAM module. We quantify the distribution alignment using Maximum Mean Discrepancy (MMD) and evaluate predictive stability via ROC-AUC during the early training phase (10 epochs). In this context, a lower MMD indicates superior robustness to perturbations, suggesting that the model successfully captures domain-invariant causal features, whereas a higher MMD signifies susceptibility to confounding variables or "style" features.

Figure \ref{fig:var_stacked} illustrates the clustering evolution of factual and "fake" (counterfactual) attention representations for three model variants. In the DCRMTA-full configuration (left), the MMD decreases significantly from 0.4780 to 0.1463. This convergence demonstrates that the synergy between Gradient Reversal Layers (GRL) and the CAM effectively aligns the factual and counterfactual latent spaces. Visually, this is manifested by the high degree of overlap between the $\mathbf{z}_f$ and $\mathbf{z}_c$ distributions, indicating the extraction of stable causal signals. Conversely, the w/o-CAM variant (right) fails to achieve meaningful distribution alignment, exhibiting persistent divergence between the two spaces. This disparity underscores the indispensable role of the Causal Attention Module in disentangling essential causal representations from latent confounders. These results provide strong empirical evidence that DCRMTA effectively eliminates confounding bias by forcing the model to learn representations that are invariant to counterfactual perturbations.

\begin{figure}[h]
  \centering
  \includegraphics[width=0.48\textwidth]{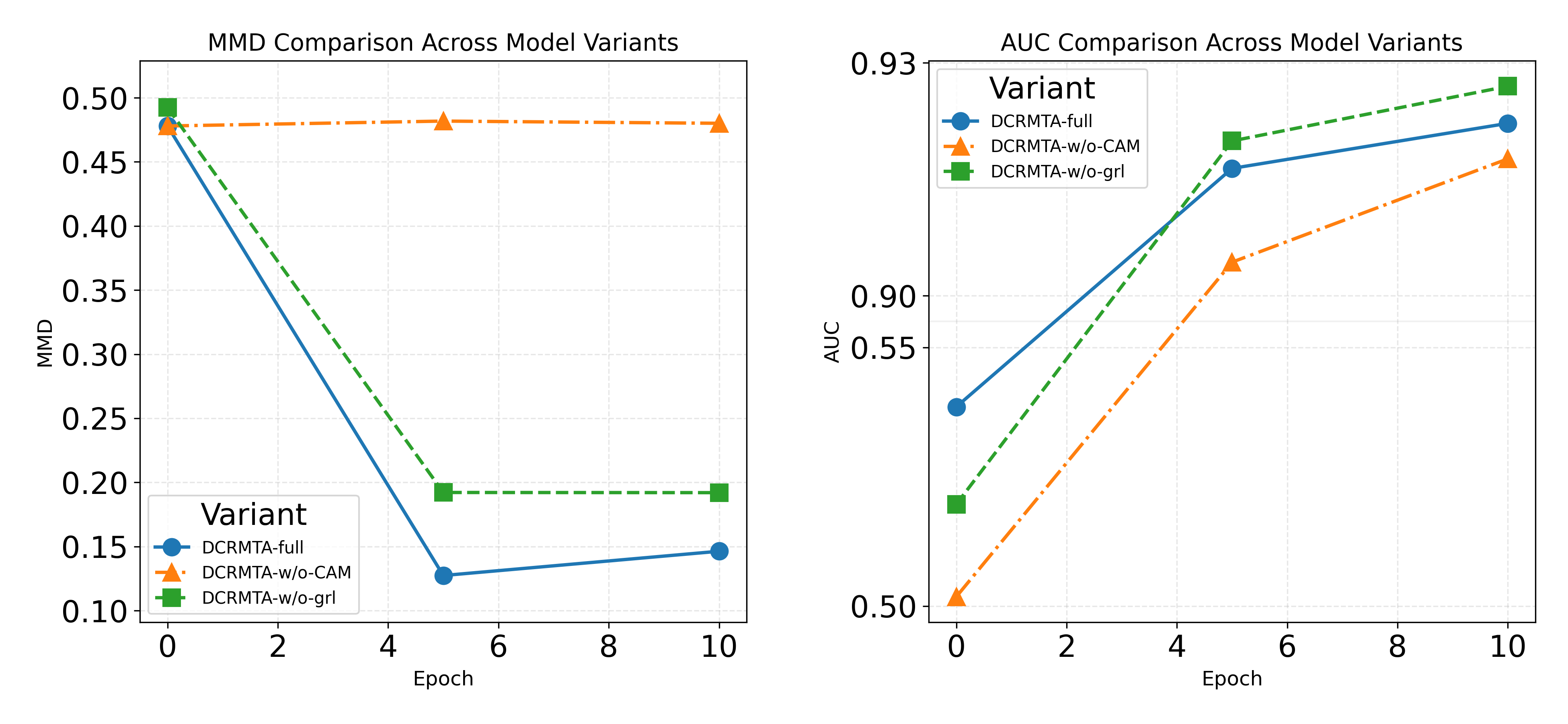}
  \caption{Model variants comparison of AUC during training}
  \label{fig:comp_variants}
\end{figure}

As illustrated in Fig.~\ref{fig:comp_variants}, the DCRMTA-full model initially exhibits a high MMD of 0.48, which precipitously declines to 0.13 by Epoch 5 and stabilizes near 0.15 by Epoch 10. This rapid convergence coincides with a substantial increase in ROC-AUC to 0.918, underscoring the synergistic role of GRL and CAM in aligning factual and counterfactual distributions while enhancing causal robustness. Interestingly, the ablation of GRL (w/o-GRL) significantly decelerates the MMD reduction trajectory—only reaching 0.19 by Epoch 5—yet yields a marginally higher final ROC-AUC. This phenomenon suggests a critical trade-off: while GRL facilitates accelerated distribution alignment, a more moderated alignment process may better preserve the discriminative structural information essential for classification. In stark contrast, the w/o-CAM variant maintains a persistently high MMD (0.48) throughout the training duration and achieves a lower ROC-AUC of 0.914. This sustained divergence indicates a fundamental inability to extract invariant causal features in the absence of the Causal Attention Module.

In essence, GRL acts as a "speed-up" factor for distribution alignment, while CAM is the structural backbone for causal disentanglement. We observe a clear trade-off: although GRL forces the model to ignore confounding features faster, excessive adversarial pressure can inadvertently prune some discriminative information, leading to a slight dip in classification. Conversely, omitting CAM leaves the model vulnerable to spurious correlations. Thus, the integration of both modules is not just a choice, but a necessity to balance the "tightness" of alignment with the "sharpness" of attribution.

\section{Conclusion and Future Work}
In this paper, we bridge a critical gap in multi-touch attribution (MTA) by systematically addressing the neglect of static user covariates in conversion modeling. We propose DCRMTA, a novel end-to-end framework that leverages causal representation learning to disentangle and integrate invariant user attributes with dynamic journey features. Through rigorous evaluation on large-scale industrial datasets (Criteo and Taobao), we demonstrate that DCRMTA consistently surpasses state-of-the-art baselines while providing more robust credit assignments by explicitly accounting for latent confounding factors. Our findings underscore the necessity of causal-aware representations in modern marketing analytics, providing a scalable solution for complex, data-sparse attribution scenarios.

Building on these results, we identify two primary avenues for future work. First, we intend to extend the framework to handle long-term conversion cycles and cross-platform user identification, where data fragmentation poses significant challenges for causal estimation. Second, we plan to explore adaptive adversarial weight scheduling to further optimize the trade-off between distribution alignment velocity and classification accuracy, thereby enhancing the model's efficiency in deployment environments with dynamic and non-stationary traffic patterns.

\input{appendix}

\bibliographystyle{IEEEtrans}
\bibliography{ref}

\end{document}

%% file: appendix.tex


\appendix

\section{Causal Representation Learning and Attention Intervention} \label{appendix: A}\subsection{Causal Structural Hypotheses}For MTA applications, we formalize the user journey through a structural causal model (SCM) based on the invariant representation learning framework \cite{mitrovic2020representation}. We assume the following three hypotheses derived from causal inference theory \cite{pearl2009causal, peters2016causal}:\begin{itemize}\item \textbf{H1 (Decomposition):} Input data $X$ is generated by a latent content variable $C$ (causal) and a style variable $S$ (confounding), such that $C \perp S$.\item \textbf{H2 (Sufficiency):} Only $C$, and not $S$, is invariant and relevant to the target conversion task $f(X)$.\item \textbf{H3 (Invariance):} Under the causal model, an intervention on $S$ (i.e., $do(S)$) does not shift the conditional distribution $P(Y|C)$:\begin{equation}\label{eq:appendix_invariance}P^{do(S=s_i)}(Y|C) = P^{do(S=s_j)}(Y|C), \quad \forall s_i, s_j \in S.\end{equation}\end{itemize}\subsection{Mechanism of Attention Intervention}To isolate the invariant causal representation $C$, we utilize the do-calculus \cite{pearl2000models} to control the attention weights. By constructing a counterfactual attention map $\bar{A}$ and applying the intervention $do(\bar{A})$, we perturb the model's focus while keeping the original feature map $X$ fixed. The Direct Causal Effect (DCE) is then estimated by the expectation of the difference between factual and counterfactual outcomes:\begin{equation} \label{eq:appendix_causal_effect}Y_{\mathrm{effect}} = \mathbb{E}_{\bar{A} \sim \mathcal{N}}\left[Y(do(A, X)) - Y(do(\bar{A}, X))\right].\end{equation}This mechanism forces the model to ignore spurious correlations encoded in $S$ and rely on the stable causal structures identified through the intervention.\subsection{Statistical Guarantees and Stability Proofs}\textit{1) Gaussian Approximation of CTR Data:}Under large-sample conditions, user click events modeled as independent Bernoulli trials $X_i \sim \text{Bern}(p)$ satisfy the De Moivre-Laplace Theorem. The empirical CTR $\hat{p} = S_n/n$ asymptotically converges to a Gaussian distribution:\begin{equation}\hat{p} \approx \mathcal{N}\left(p,, \frac{p(1-p)}{n}\right).\end{equation}This justifies the use of Gaussian-based perturbations ($\Sigma_F$) for modeling attention interventions in high-traffic MTA environments.

\textit{2) Additivity and Distributional Integrity:}Let the factual attention $A \sim \mathcal{N}(\mu_A, \Sigma_A)$ and a synthetic counterfactual noise $F \sim \mathcal{N}(0, \Sigma_F)$ be independent. The perturbed attention $A' = A + F$ satisfies:\begin{equation}A' \sim \mathcal{N}\left(\mu_A,, \Sigma_A + \Sigma_F\right).\tag{Additivity Property}\end{equation}By ensuring $\Sigma_F \ll \Sigma_A$, $A'$ remains within the original semantic manifold, preventing the "catastrophic forgetting" of content during the intervention process.

\textit{3) Stability via Layer Normalization:}Given content features $C = WA + b$, the injected noise $F$ results in $C' = C + WF$. Since $\mathbb{E}[F] = 0$, the feature expectation remains invariant ($\mathbb{E}[C'] = \mathbb{E}[C]$), though variance increases by $W\Sigma_F W^\top$. To suppress this variance shift, Layer Normalization (LN) is applied:\begin{equation}\text{LN}(C') = \gamma \cdot \frac{C' - \mu_{C'}}{\sigma_{C'}} + \beta.\end{equation}The LN operation acts as a projection that realigns the perturbed features with the learned distribution $\mathcal{N}(\beta, \gamma^2)$, thereby preserving the underlying content semantics while enabling the estimation of causal effects through $do(A+F)$.